\documentclass[runningheads]{llncs}
\usepackage[T1]{fontenc}
\usepackage{graphicx}
\usepackage{pdfpages}

\usepackage{longtable}
\usepackage{makecell}

\usepackage{amsmath}

\begin{document}

\includepdf[pages={1}]{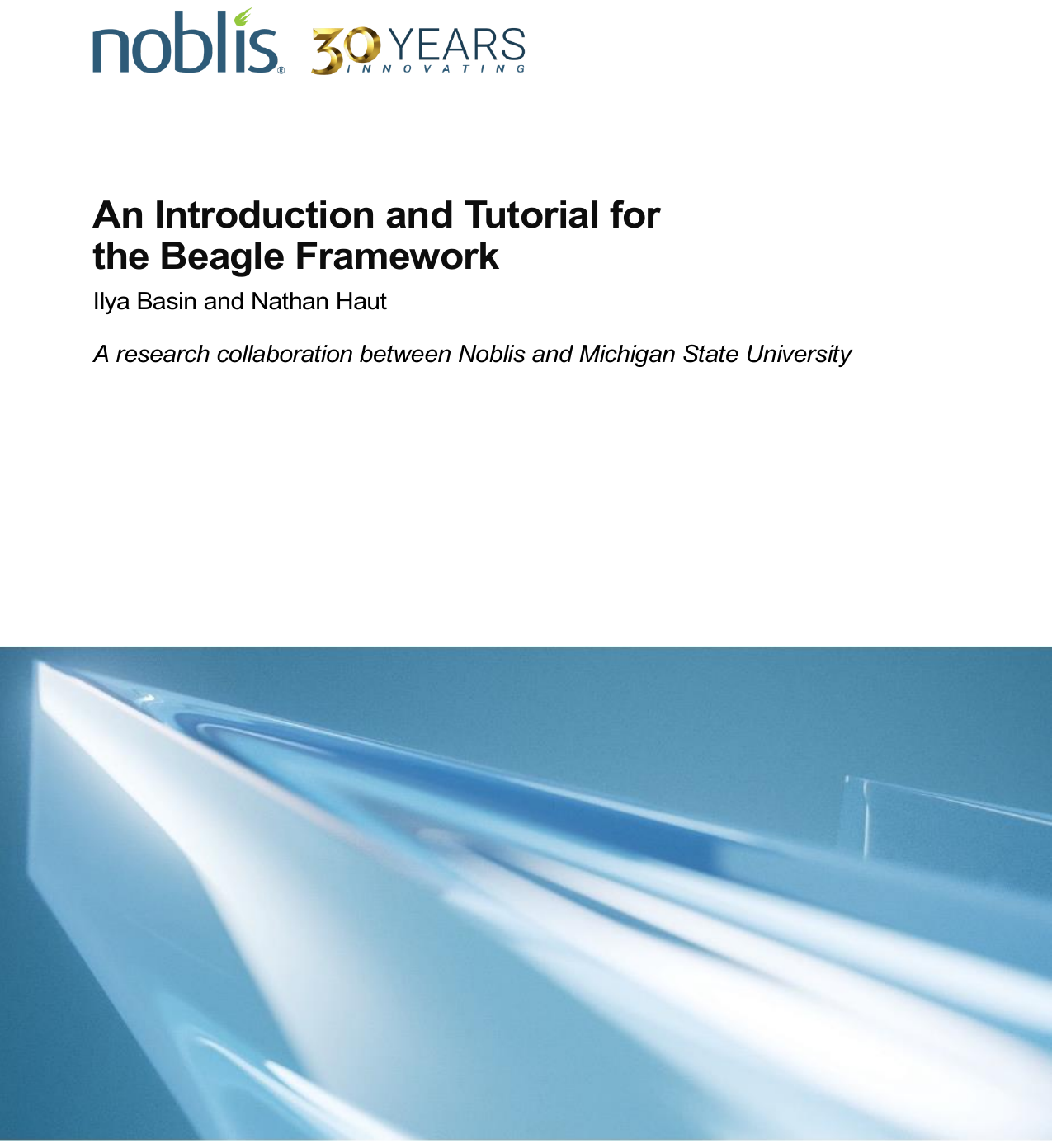}
\newpage

\title{An Introduction and Tutorial for the Beagle Framework}
\titlerunning{Beagle Tutorial}
%
\author{Ilya Basin\inst{2}
\and
Nathan Haut\inst{1,2}
}
\authorrunning{}
%
\institute{Michigan State University, East Lansing MI 48824, USA\\
\and
Noblis, Reston VA , USA,
\email{Ilya.Basin@noblis.org}}
\maketitle              
\begin{abstract}

The Beagle framework is a GPU-based genetic programming framework that enables highly efficient genetic programming search using large population sizes by leveraging NVIDIA GPUs. This technical guide provides an introduction to the Beagle framework and provides detailed instructions for using the framework for symbolic regression problems.


\end{abstract}
%


%

\section{Introduction}
Beagle is an open-source symbolic regression framework designed with NVIDIA GPU acceleration as a primary objective. Beagle was created by Ilya Basin as an internally funded R\&D project at Noblis, a non-profit science and technology organization headquartered in the Washington, D.C. metro area.

Beagle was specifically designed to optimize scaling to large population sizes by taking advantage of heterogeneous computing environments to efficiently explore large/complicated search spaces \cite{beagle}. A benchmarking study demonstrated that the Beagle GPU-accelerated genetic programming framework achieves substantially superior performance compared to existing CPU-based frameworks when evaluated on the Feynman100 benchmark suite, a widely adopted standard for assessing symbolic regression capabilities in the field \cite{haut2026gpu}. 

The Beagle source code can be accessed here: \\
\url{https://github.com/Noblis/beagle-v1.x} 

The goal of this technical tutorial is to first discuss and introduce an overview of the framework (Section \ref{sec:Overview}) and then to describe key technical details of using Beagle with step-by-step guides to install and get started using the Beagle framework (Section \ref{sec:using}). 

\section{Framework Overview}\label{sec:Overview}

In this section, we discuss the key framework details and design choices that make Beagle a highly efficient GPU Genetic Programming framework. 

\subsection{Software Environment}

Beagle utilizes the ILGPU open-source C\# library. ILGPU allows low-level GPU programming rather than relying on high-level GPU libraries. This allowed for the design of a completely custom CUDA kernel to best utilize the hardware. Beagle's development stack is .Net 10, which is cross-platform, ensuring robust performance on both Linux and Windows operating systems. Since Mac's do not natively host NVIDIA GPUs, GPU acceleration in Beagle on a Mac is currently not possible. Apple's custom GPUs and the Metal framework are becoming quite powerful, so depending on the future use-cases of Beagle, it may be worthwhile to explore that extension. 

\subsection{Hardware Optimized}

Beagle is designed to automatically distribute work to as many GPUs and CPUs as are available, thus making it increasingly powerful as hardware access scales. For users with access to HPC systems, Beagle could be used to effectively explore massive search spaces. 

On the other end, Beagle is also optimized to work on modest hardware with smaller GPU compute and memory capacity. In this case, Beagle manages populations in batches to ensure the GPU memory is never exceeded. It is also possible to run Beagle entirely on CPU (using ILGPU GPU emulation feature), although running only on a CPU is much slower and should be reserved for debugging purposes only. 

\subsection{Beagle Taxonomy}

While it is now realized that the Beagle framework is essentially a Genetic Programming system, it was developed independent of the field of Genetic Programming and relied only on biological inspiration of evolution. As a result, the naming conventions used in the Beagle software/API are different than what is used in the general field of Genetic Programming. For the concepts that have a functional equivalent in the field of Genetic Programming, we've provided Table \ref{tab:terms} which matches the internal Beagle framework terms to those commonly used in GP. In this guide, we try to adhere to GP conventional terminology where possible, but for those interested in looking at the source code, this table will be useful. 

\begin{table}[h]
\centering
\caption{Comparison of GP conventional terms versus Noblis-originated Beagle terminology.}
\label{tab:terms}
\begin{tabular}{|c|c|}
\hline
{GP Standard Term} & {Beagle Term}   \\ \hline \hline
{  Individual}             & {  Organism}      \\ \hline
{  Population}             & {  Colony}        \\ \hline
{  Evolutionary Run}       & {  Experiment}    \\ \hline
{  Parsimony Pressure}     & {  Tax on Length} \\ \hline
{  Fitness Evaluation}     & {  Scoring}       \\ \hline
{  Fitness}                & {  Accuracy}     \\ \hline
\end{tabular}
\end{table}

\subsection{Dividing Tasks Between CPU and GPU}

In Beagle's architecture, the GPU manages model and fitness function evaluations, while the CPU handles the genetic operations (mutation, initialization, etc.). The current implementation does not employ crossover. To reduce the overhead from data transfer between GPU and CPU, which can be substantial, Beagle evaluates multiple fitness cases per individual within a single generation. This avoids sending individuals back and forth between the CPU and GPU since the individual is sent to the GPU once and the aggregated fitness value is returned for the full batch of fitness cases. While the batch size is configurable, Beagle typically uses batch sizes of 512 or 1024 to best align with the hardware. Within a single generation with a population size of 1 million individuals and a batch size of 512 this means Beagle would conduct 512 million total fitness case evaluations. Within a single generation and if the GPU memory is large enough, all genomes in the population are transmitted to the GPU simultaneously, with each genome evaluating a batch of fitness cases. Upon completion, the aggregated fitness results for all individuals are returned to the CPU.

Since NVIDA GPUs have a limit of maximum 1024 threads per block, we cap the number of fitness cases at 1024 per generation to ensure one model is only ever present in one block. While it would technically be possible to distribute models over multiple blocks, this could slow down fitness aggregation since aggregating through global memory would be much slower than aggregating over block shared memory. As well, it would increase the memory footprint which for large populations or small GPU memory sizes could force batching which would also slow down performance. 

\subsection{Avoiding Thread Divergence}

While for the most part, thread divergence on parallel CPU threads is of little concern, thread divergence on GPUs can be disastrous if it regularly occurs within within the same warps. This is because CUDA threads within a warp run in lockstep, so if one thread goes down one branch and the other thread down the other branch, both threads must execute the code for both branches, thus approximately doubling the runtime. 

To mitigate this, during GPU kernel execution, each CUDA block corresponds to a single individual (meaning the number of blocks equals the population size), and each CUDA thread within a block processes a single fitness case for that individual. Since only one model is ever operating within a single warp, occurrences of thread divergence should be much less common than if we had mixed models within the same warp. For some search spaces, thread divergence is even eliminated entirely using this setup. 

\subsection{CPU Optimizations}

On the CPU side, Beagle's implementation carefully manages memory and nearly eliminates garbage collection, memory fragmentation, and memory allocation/ deallocation overhead, which can otherwise consume substantial computational resources in evolutionary algorithms. Typically, evolutionary algorithms must continuously allocate and deallocate memory as individuals are born or die, causing excessive garbage collection activity and memory fragmentation that can significantly degrade performance. To address this issue, Beagle never deallocates memory for individuals. Instead, memory from deceased individuals is placed into a "dead pool" and reused when memory is required for new individuals. Generally, Beagle avoids deallocating memory except for C\# strings, thereby minimizing garbage collection and memory defragmentation overheads

\subsection{Representation}

Beagle employs a custom-developed Genome Computer Language (GCL), which is a Linear Genetic Programming (LGP) language built on Reverse Polish Notation (RPN). Using LGP instead of the conventional tree-based approach makes GCL more compatible with garbage collection and GPU operations. The requirement that GPU memory be allocated statically (e.g., no heap and no stack) was one of the reasons that necessitated the choice of LGP architecture, since tree-based data structures commonly employed in GP did not lend themselves well to the GPU programming model.

\subsection{Always Viable Mutations}

While most mutations produce a viable genome (individual runs without throwing errors), in the event an invalid genome is produced, a random correction mechanism is applied to ensure the offspring is functional. This helps ensure that model evaluation compute time is spent evaluating only functional individuals.

\subsection{Fitness Functions}

Beagle, by default, uses a correlation fitness function \cite{Haut2023,keijzer2004scaled} since it has been shown to provide performance benefits, even when used in GPU environments \cite{haut2026gpu}. However, Beagle can support any user-defined fitness function that performs a point-to-point comparison between the model output vector and the target data vector. 

The core of the correlation fitness function is shown in Equation \ref{eq:r1}, 


\begin{equation}\label{eq:r1}
    r = \frac{\sum_{i=1}^{N} (y_i - \bar{y}) \, (f(x_i) - \overline{f(x)})}
{\sqrt{\sum_{i=1}^{N} (y_i - \bar{y})^2 \; \sum_{i=1}^{N} (f(x_i) - \overline{f(x)})^2}}
\end{equation}
where $N$ represents the number of data points $i$, $y_i$ denotes the target output, $f(x_i)$ represents the model's output given input data $x_i$, $\bar{y}$ represents the mean target output, and $\overline{f(x)}$ represents the mean model's output. The $r$ value is then squared. 

When a model produces invalid numbers (NaN, Inf, etc.) or a dataset contains them, Beagle does not remove these values but instead handles those point pairs separately by rewarding models that have invalid values in the same locations as the target dataset. Therefore, the $r$ value only reflects point pairs where both the model and target output contain valid numbers. After obtaining $r^2$, we then calculate the total score using Equation \ref{eq:score1}, 

\begin{equation}\label{eq:score1}
    \text{score} = M \, r^4 \left( N - (c_1 + c_2) \right) - M (c_1 - c_2)
\end{equation}
where $N$ is the total number of fitness cases, $M$ is a maximum score parameter configured within Beagle, $c_1$ is the count of valid/invalid pairs, and $c_2$ is the count of invalid/invalid pairs. 

\subsection{Monte-Carlo Ranking}

Since GPU accelerations allows Beagle to scale to populations sizes in the millions, sorting the whole population for a rank-based selection becomes computationally very expensive. Rather than ranking the whole population each generation, Beagle takes a random sample of 100 individuals and uses those 100 individuals to estimate the population percentiles. Through one pass of the population then, individuals are assigned to their specific percentile and are given rank proportional breeding rights. Specifically, the assigned breeding rights per group result in individuals from each group having an average number of offspring as is outlined in Table \ref{tab:percentile_children}.

\begin{table}[h]
\centering
\caption{Expected Number of Children by Percentile}
\label{tab:percentile_children}
\begin{tabular}{|c|c|}
\hline
{Percentile} & {Expected number of children} \\ \hline \hline
{ $\geq 90$ } & { 4 }    \\ \hline
{ $\geq 80$ } & { 2.5 }  \\ \hline
{ $\geq 70$ } & { 1.5 }  \\ \hline
{ $\geq 60$ } & { 1 }    \\ \hline
{ $\geq 50$ } & { 0.5 }  \\ \hline
{ $\geq 40$ } & { 0.25 } \\ \hline
{ $\geq 30$ } & { 0.15 } \\ \hline
{ $\geq 20$ } & { 0.06 } \\ \hline
{ $\geq 10$ } & { 0.03 } \\ \hline
{ $\geq 0$ }  & { 0.01 } \\ \hline
\end{tabular}
\end{table}

\subsection{Population Size Dynamics}

The target population size in any generation is not a hard value, rather it is the average population size that will occur over the length of evolution. Due to the way that the offspring are produced, the population size tends to vary within +/- 20\% of the target population size in any generation. 

Beagle can support a static target population size where the goal population size across all generations is the same. Beagle can also support dynamic target population sizes where the target population size changes as a function of the generation number. Often, we have found there to be benefits to start with a large population size and over generations shrink the population size \cite{haut2026effects}. This promotes exploration early and then transitions to exploitation once promising regions of the search space are identified. 


\section{Using the Beagle Framework}\label{sec:using}

This section provides a comprehensive guide to the Beagle framework. We present the installation procedure (Section \ref{sec:install}), describe the Genome Computer Language (GCL) employed by Beagle for genetic programming operations (Section \ref{sec:gcl}), provide guidance for interpreting Beagle output (Section \ref{sec:output}) and interacting with Beagle during a run (Section \ref{sec:interact}), list the available command-line options (Section \ref{sec:command}), describe the configuration parameters for custom machine learning experiments (Sections \ref{sec:configExperimentMand}-\ref{sec:ConfigCustFit}), and describe useful command-line utilities (Section \ref{sec:ConfigUtils}). Taken together, Section \ref{sec:using} serves as a technical reference for using the Beagle framework.

\subsection{Installation}\label{sec:install}

Beagle requires a computer system equipped with at least one NVIDIA GPU. Any CUDA-compatible NVIDIA GPU is sufficient, although higher performance is achieved with GPUs possessing greater memory capacity and more recent architectures. The Beagle framework automatically detects and supports multi-GPU configurations, enabling scalable deployment across diverse hardware environments ranging from high-performance computing servers with multiple GPUs to individual workstations or laptops with a single GPU.

The Beagle framework is implemented in C\#/.NET 10 utilizing the open-source ILGPU library, which provides CUDA-like functionality within the C\# ecosystem. This architecture ensures cross-platform compatibility across Windows and Linux operating systems. macOS is currently unsupported due to the absence of NVIDIA GPU hardware in Apple systems. Although the .NET framework can be executed via command-line tools provided with the .NET SDK, development and code modification may be facilitated through .NET-compatible integrated development environments (IDEs), such as Visual Studio, Visual Studio Code, or JetBrains Rider. 
While the screenshots presented in this paper were captured using Visual Studio 2026, other compatible IDEs provide equivalent functionality.

In addition to .NET 10, Beagle requires that a CUDA driver be installed. If a machine has an NVIDIA GPU, it is very likely that a CUDA driver is already installed, in which case, nothing needs to be done. 

If running Beagle with \texttt{UseLibDevice}, it is also necessary to install the CUDA Toolkit. By default \texttt{UseLibDevice} is disabled since benchmarking has shown worse performance when enabled. Therefore, for most users, the CUDA Toolkit is not needed. The \texttt{useLibDevice} flag is discussed further in Section \ref{sec:command}. The CUDA Toolkit typically includes the necessary NVIDIA driver, eliminating the need for a separate driver installation. The toolkit is available for free download at 
\texttt{https://developer.nvidia.com/cuda-downloads}. 

Note, at the time of writing, there is a known compatibility issue between CUDA 13.3 and ILGPU. Specifically, the nvvm64\_40\_0.dll file has been moved into the directory "Program Files/NVIDIA GPU Computing Toolkit/CUDA/\allowbreak v13.3/nvvm/bin/x64" but is expected one level up in the "bin" folder. If you encounter a missing dll error (\texttt{Unhandled exception. System.NotSupportedExc-\allowbreak eption: Could not find NVVM DLL in 'C:\textbackslash Program Files\textbackslash NVIDIA GPU C- \allowbreak omputing  Toolkit\textbackslash CUDA\textbackslash v13.3\textbackslash nvvm\textbackslash bin'}), just copy the file one directory up into the "bin" to fix the issue. 

Beagle is an open-source framework distributed under Apache License 2.0. The source code repository is publicly available at \texttt{https://github.com/Noblis\allowbreak /beagle-v1.x}.

\begin{figure}
    \centering
    \includegraphics[width=0.9\linewidth]{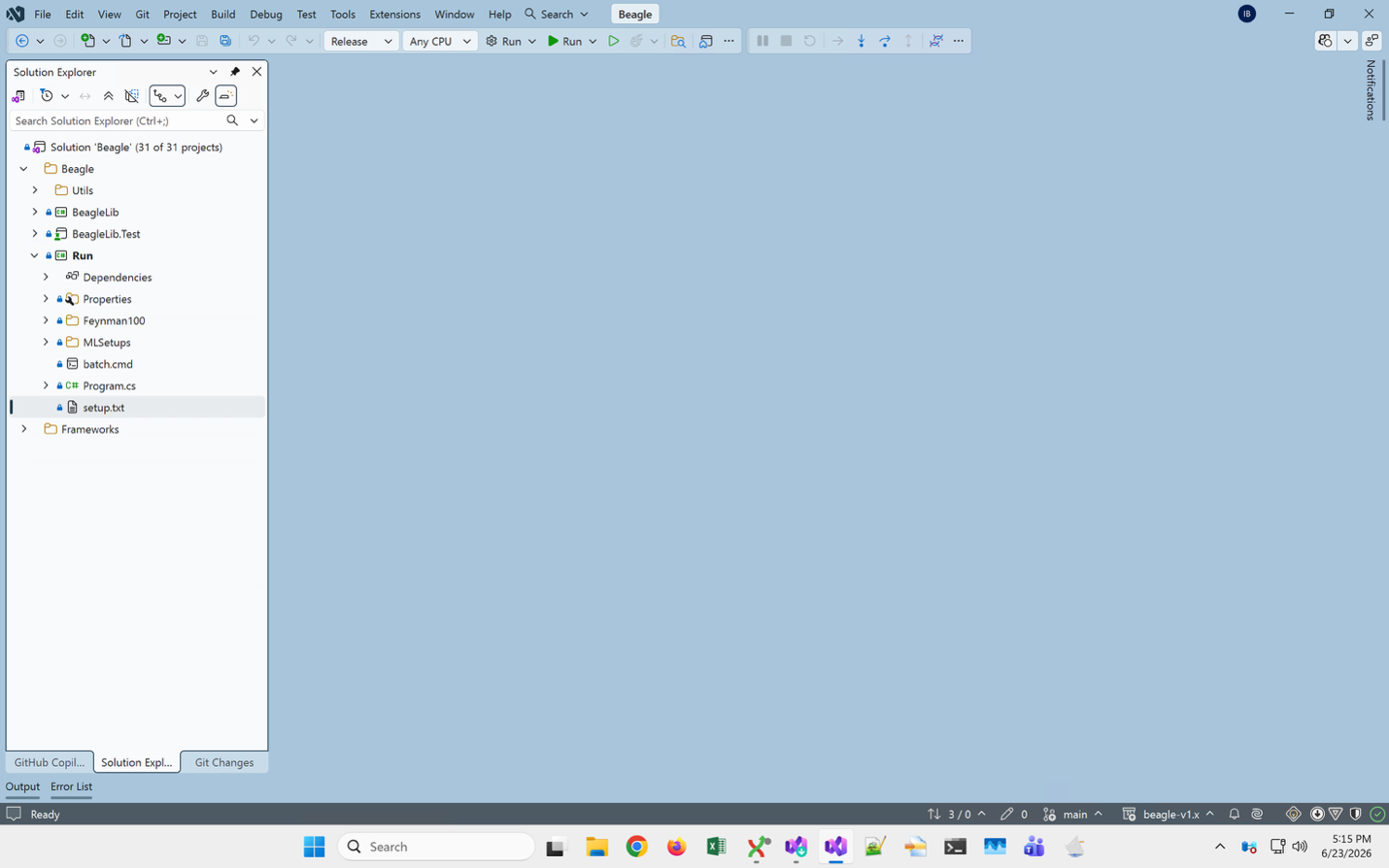}
    \caption{
     Visual Studio 2026 with Beagle solution loaded 
    }
    \label{fig:vsBeagle}
\end{figure}

Select the Run project in Beagle solution as a startup project and run it. The output should look like Figure 2.

\begin{figure}
    \centering
    \includegraphics[width=0.9\linewidth]{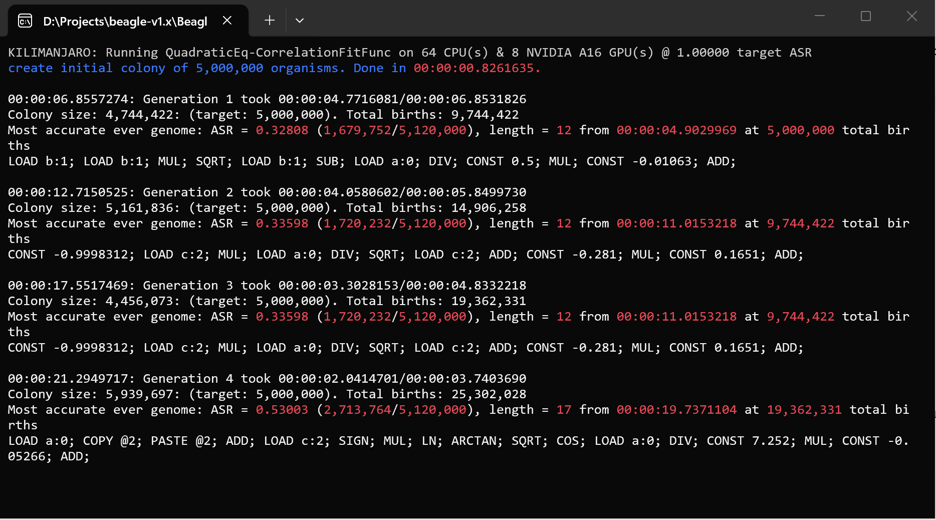}
    \caption{
     An example of a Beagle machine learning run 
    }
    \label{fig:example}
\end{figure}

\subsection{Genome Computer Langauge (GCL)}\label{sec:gcl}

\begin{figure}
    \centering
    \includegraphics[width=0.9\linewidth]{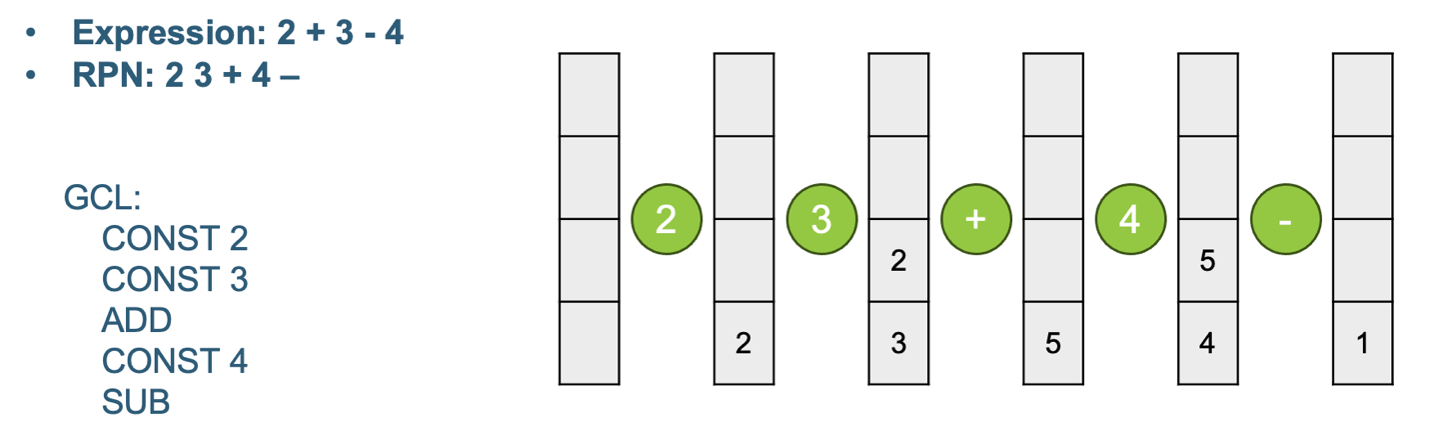}
    \caption{
     A Mathematical expression in traditional form, RPN form, and represented as GCL. 
    }
    \label{fig:GCL}
\end{figure}

Beagle employs a bespoke Linear Genetic Programming (LGP) language called Genome Computer Language (GCL), which utilizes Reverse Polish Notation (RPN). RPN employs a stack-based architecture for operand management (Figure \ref{fig:GCL}). Unary operations (e.g., squaring) operate on the top stack element and replace it with the computational result. Binary operations (e.g., adding) operate on the top two stack elements and replace both with the resulting value.

\begin{figure}
    \centering
    \includegraphics[width=0.9\linewidth]{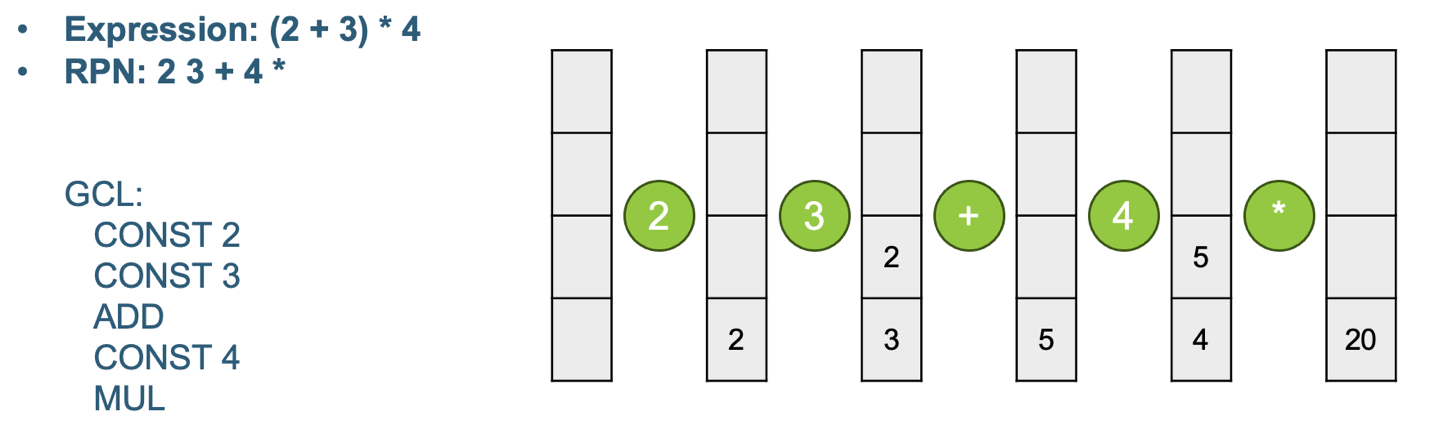}
    \caption{
     A Mathematical expression with parentheses in traditional form, RPN form, and represented as GCL.
    }
    \label{fig:RPN}
\end{figure}

A key advantage of RPN is the elimination of parentheses in mathematical expressions (Figure \ref{fig:RPN}). 

While RPN theoretically supports an unbounded stack size, allowing arbitrary operand accumulation prior to operation execution, Beagle's implementation restricts the stack to 32 elements. This parameter can be modified by adjusting the \textbf{StackSize} constant in the BConfig.cs file within the BeagleLib project. The Figure \ref{fig:Stack} below shows an example of using more than two operands in the stack at a time.

\begin{figure}
    \centering
    \includegraphics[width=0.9\linewidth]{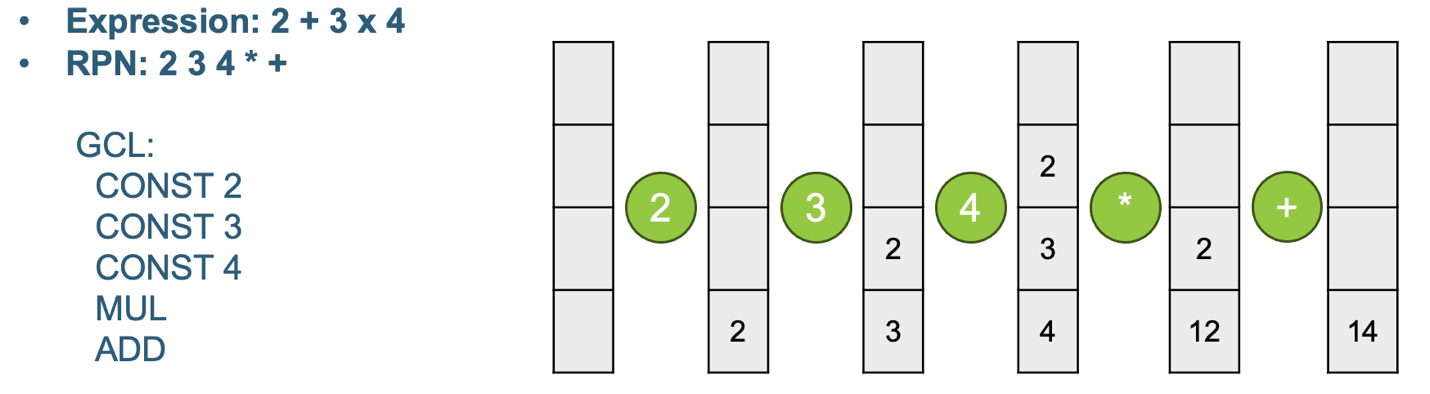}
    \caption{
     Utilization of the RPN stack for storing multiple operands simultaneously.
    }
    \label{fig:Stack}
\end{figure}

Table \ref{tab:commands} presents the complete set of GCL operations. Note that during machine learning experiments, individual operations can be selectively disabled, constraining Beagle to generate solutions using only a specified subset of the GCL instruction set. See section \ref{sec:ConfigExperimentOpt} for instructions on how to exclude certain GCL commands from a given machine learning run.

\begin{longtable}{|c|p{7cm}|c|}
\caption{GCL commands}
\label{tab:commands}\\
\hline
GCL Command & Operation & Example \\ \hline \endhead
\hline
    CONST & Loads a floating-point number onto the stack. The example loads $\pi$ onto the stack. & CONST 3.14 \\ \hline
    LOAD & Loads a floating-point input parameter onto a stack. The example loads input parameter a, which also happens to be the first input parameter onto a stack. Beagle input parameters are zero-base indexed. & LOAD a:0 \\ \hline
    DUP & Pushes a duplicate value from the top of the stack onto the stack. After the operation is completed, the two top-most stack parameters will be identical. & DUP \\ \hline
    SWAP & Swaps two top-most elements on the stack. & SWAP \\ \hline
    DEL &  Deletes the top-most stack element, effectively POPing the stack. & DEL \\ \hline
    ADD & Adds two top-most stack elements, removes them both from the stack and pushes the result onto the stack. & ADD \\ \hline
    SUB & Subtracts top-most stack element from the second-top-most stack element, removes them both from the stack, and pushes the result onto the stack. & SUB \\ \hline
    MUL & Multiplies two top-most stack elements, removes them both from the stack, and pushes the result onto the stack. & MUL \\ \hline
    DIV & Divides second-top-most stack element by the top-most stack element, removes them both from the stack, and pushes the result onto the stack. & DIV \\ \hline
    POW & Raises the second-top-most stack element to the power of the top-most stack element, removes them both from the stack, and pushes the result onto the stack. & POW \\ \hline
    SIGN & Changes the sign of the top-most stack element. & SIGN \\ \hline
    SQUARE & Squares the top-most element on the stack. & SQUARE \\ \hline
    CUBE & Cubes the top-most element on the stack. & CUBE \\ \hline
    SQRT &  Takes a square root of the top-most element on the stack. & SQRT \\ \hline
    CBRT & Takes a cubic root of the top-most element on the stack. & CBRT \\ \hline
    LN & Takes a natural logarithm of the top-most element on the stack. & LN \\ \hline
    SIN & Takes a sine of the top-most element on the stack. & SIN \\ \hline
    COS & Takes a cosine of the top-most element on the stack. & COS \\ \hline
    TAN & Takes a tangent of the top-most element on the stack. & TAN \\ \hline
    ARCSIN & Takes an arcsine of the top-most element on the stack. & ARCSIN \\ \hline
    ARCCOS & Takes an arccosine of the top-most element on the stack. & ARCCOS \\ \hline
    ARCTAN & Takes an arctan of the top-most element on the stack. & ARCTAN \\ \hline
    TANH & Takes a hyperbolic tangent of the top-most element on the stack. & TANH \\ \hline
    EXP & Takes an exponential function (ex) of the top-most element on the stack. & EXP \\ \hline
    COPY, PASTE & The COPY and PASTE operations constitute GCL's variable implementation mechanism. These operations function as a paired command structure: in valid GCL programs, they must occur in corresponding pairs. The numeric identifier (e.g., @1, @2, @3) serves as the variable reference, with @2 representing such an identifier in the given example. Note that each Copy operation corresponds to exactly one Paste operation, establishing a one-to-one mapping.

    \vspace{2mm}
    
    The COPY operation copies the top-most value from the stack onto the variable. Note that this operation does not make any changes to the stack. The PASTE operation takes the value from the variable and pushes it onto the stack. After the operation is completed, the variable is destroyed. & \makecell[t]{ COPY @2  \\ ... \\ ... \\ PASTE @2} \\ \hline
\end{longtable}

\subsection{Understanding Beagle Output}\label{sec:output}

\begin{figure}
    \centering
    \includegraphics[width=0.9\linewidth]{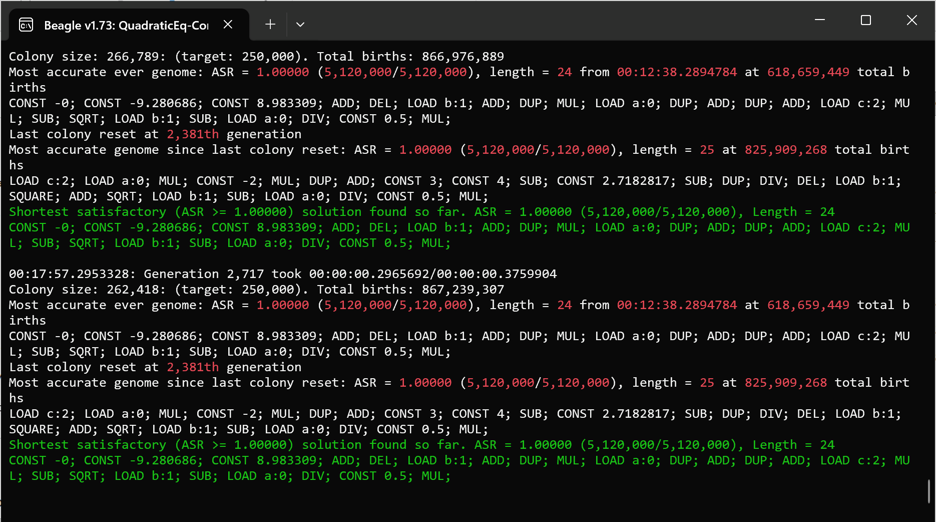}
    \caption{
    Sample console output of a Beagle run.
    }
    \label{fig:Output}
\end{figure}

Figure \ref{fig:Output} represents a sample console output of a Beagle run. Note that, in addition to being displayed on the console, Beagle output is also stored in a .txt file \emph{AppOutput} directory, which is a subdirectory in the working folder of the executable. For a Windows machine, this would be the file path: \texttt{beagle-v1.x/\allowbreak Beagle/Run/bin/Release/net10.0/AppOutput}.

Let’s review the output for Generation 2,717. 
Line 1 (\texttt{00:17:57.2953328: Generation 2,717 took 00:00:00.2965692/00:00:00.3759904}) displays the elapsed time since the start of the run, followed by the generation number and the computational time required for that specific generation. The numerator represents the time spent exclusively in GPU kernel execution, while the denominator indicates the total generation time encompassing both CPU and GPU operations.

Line 2 (\texttt{Colony size: 262,418: (target: 250,000). Total births: \\ 867,239,307}) displays the current population size, the target population size, and the cumulative number of births since the start of the run.

Line 3 (\texttt{Most accurate ever genome: ASR = 1.00000 (5,120,000/\\5,120,000), length = 24 from 00:12:38.2894784 at 618,659,449 total\\ births}) presents information regarding the most accurate genome identified during the run, quantified by the Average Score Ratio (ASR). The ASR is defined as the ratio of cumulative rewards obtained across all evolutionary runs for a given genome to the theoretical maximum achievable rewards. An ASR value of 1.0 indicates a “perfect” solution. Additional metrics include the raw reward values versus raw maximum possible rewards, genome length (i.e., the number of GCL commands), and temporal metadata indicating both elapsed time and the cumulative birth count at the time of the genome’s creation.

Line 4 (\texttt{CONST -0; CONST -9.280686; CONST 8.983309; ADD; DEL; LOAD b:1; ADD; DUP; MUL; LOAD a:0; DUP; ADD; DUP; ADD; LOAD c:2; MUL; SUB; SQRT; LOAD b:1; SUB; LOAD a:0; DIV; CONST 0.5; MUL;}) displays the actual GCL code for the “most accurate ever genome.”

Line 5 (\texttt{Last colony reset at 2,381th generation}) warrants additional explanation. Beagle implements a mechanism to mitigate prolonged convergence on suboptimal local minima. The MLSetup configuration (detailed in Section \ref{sec:ConfigExperimentOpt}) includes the parameter \textbf{TotalBirthsToResetColonyIfNoProgress}, which defines the threshold number of births without improvement needed to trigger a population reset. When this threshold is reached, the population is reinitialized while maintaining a record of the globally optimal genome, which is excluded from the new population's genetic pool to promote exploration of alternative solution spaces.

Line 6 (\texttt{Most accurate genome since last colony reset: ASR = 1.000\\00 (5,120,000/5,120,000), length = 25 at 825,909,268 total births})\\ displays the information on the most accurate genome since the last population reset.

Line 7 (\texttt{LOAD c:2; LOAD a:0; MUL; CONST -2; MUL; DUP; ADD; CONST 3; CONST 4; SUB; CONST 2.7182817; SUB; DUP; DIV; DEL; LOAD b:1; SQUARE; ADD; SQRT; LOAD b:1; SUB; LOAD a:0; DIV; CONST 0.5; MUL;}) displays the actual GCL code for the most accurate genome since the population was reset.

Line 8 (\texttt{Shortest satisfactory (ASR >= 1.00000) solution found so far. ASR = 1.00000 (5,120,000/5,120,000), Length = 24}) displays information on the shortest genome that meets the minimum acceptable performance threshold. While Line 4 presents the most accurate solution, the MLSetup configuration (detailed in Section \ref{sec:ConfigExperimentOpt}) allows a user to specify a minimum acceptable ASR through the \textbf{SolutionFoundASRThreshold} parameter, which may be set below 1.0. The genome reported on Line 8 represents the most compact solution satisfying this minimum ASR criterion. In the example shown, the solution achieves optimal performance.

Line 9 (\texttt{CONST -0; CONST -9.280686; CONST 8.983309; ADD; DEL; LOAD b:1; ADD; DUP; MUL; LOAD a:0; DUP; ADD; DUP; ADD; LOAD c:2; MUL; \\SUB; SQRT; LOAD b:1; SUB; LOAD a:0; DIV; CONST 0.5; MUL;}) displays the actual GCL code for the shortest satisfactory solution.

\subsection{Interacting with Beagle During a Run}\label{sec:interact}

\begin{figure}
    \centering
    \includegraphics[width=0.9\linewidth]{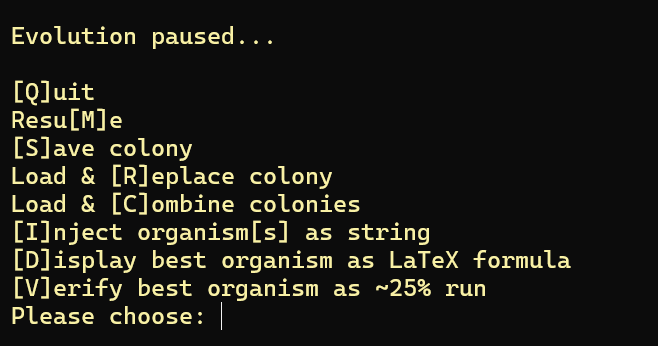}
    \caption{
    Menu for interacting with Beagle during a run
    }
    \label{fig:Interact}
\end{figure}

At any time during a run, a user can press Esc to pause the evolution and display options that allow them to interact with the population (see Figure \ref{fig:Interact}).

\begin{table}[h]
\centering
\caption{Beagle interactive commands}
\label{tab:interact}
\begin{tabular}{|c|p{6cm}|}
\hline

    Command & Effect \\ \hline
    Q   & Quits the run \\ \hline
    M   & Resumes the run \\ \hline
    S   & Saves the current population in a file \\ \hline
    R   & Loads a population from a file, replacing the current population \\ \hline
    C   & Loads a population from a file, adding to the current population \\ \hline
    I   & Adds one or more genomes to the population by copying and pasting the genome's GCL \\ \hline
    D   & Opens \texttt{https://arachnoid.com/} and displays the “most accurate ever genome” as a mathematical formula \\ \hline
    V   & Runs the “most accurate ever genome” against a brand-new testing data set (1/4 of the size of a normal batch set, defined by \textbf{ExperimentsPerGeneration} in Section \ref{sec:ConfigExperimentOpt} below). A genome is classified as providing a “perfect” solution when the discrepancy between predicted and actual values for each individual test case does not exceed 0.1\%. \\ \hline
    
\end{tabular}
\end{table}

\subsection{Command-Line Parameters}\label{sec:command}

Beagle supports execution via optional command-line parameters. Table \ref{tab:command} presents the available command-line parameters, which are case-insensitive, should be separated with spaces, are not all required to be present, and may be specified in any order. These command-line parameters are typically used for executing multiple runs in batch files.

The command-line parameters can be useful for running the Feynman100 benchmark \cite{data}. The code to run the benchmark is integrated into Beagle, such that symbolic regression for each of the 100 formulas make up the Feynman100 benchmark can be invoked using the \textbf{RunFeynman=\{1-100\}} command-line parameter described below in Table \ref{tab:command}.

\begin{table}[h]
\centering
\caption{Beagle command-line parameters}
\label{tab:command}
\begin{tabular}{|c|p{6cm}|}
\hline

    Command-Line Parameter & Effect \\ \hline
    NoEscMenu   & Disables user interactivity (see Section \ref{sec:interact}) by directing Beagle to suspend keyboard monitoring. This parameter is essential for automated scheduled executions operating in non-interactive environments. When this option is not specified, attempts to access keyboard input may result in application termination. \\ \hline
    StopAfterMin=\{minutes\} & Directs Beagle to terminate execution after a specified number of minutes. Upon reaching this time limit, validation and display of the “most accurate ever genome” is performed (analogously to the V and D commands described in Table 2\ref{}) \\ \hline
    RunFeynman=\{1-100\} & Directs Beagle to execute symbolic regression on a specific formula from the Feynman100 benchmark suite (index range: 1-100). \\ \hline
    UseLibDevice & Forces GPU to use custom mathematical operators. Benchmarking hasn't shown this to be beneficial. \\ \hline
\end{tabular}
\end{table}

For instance, the parameter configuration \textbf{StopAfterMin=30 RunFeynman=11 NoEscMenu} initiates a 30-minute symbolic regression execution on Feynman100 formula 11 under the non-interactive mode. 

Command-line commenting is supported via the \textbf{\#} delimiter, which causes all subsequent text to be treated as a comment. As an example, \textbf{StopAfterMin=30 RunFeynman=11 \#NoEscMenu} performs the same 30-minute execution of Feynman100 formula 11 while maintaining interactive mode, as the \textbf{NoEscMenu} parameter is effectively commented out.

\subsection{Configuration Parameters for Custom Machine Learning Experiments: Subclassing MLSetup - mandatory overrides}\label{sec:configExperimentMand}

To set up a brand-new ML run, users must subclass the MLSetup base class and override several methods and properties. While the MLSetup subclass can be placed anywhere within the Run project, by convention, it is placed in the MLSetups directory under the Run project.
The following two methods are declared abstract in the base class and therefore must be overridden:

\begin{verbatim}
    public abstract string[] GetInputLabels();
    
    public abstract (float[], float)
        GetNextInputsAndCorrectOutput(float[] inputsToFill);

\end{verbatim}

The \textbf{GetInputLabels} method must return an array of strings representing labels for input variables (e.g., ["x," "y"] or ["a," "b," "c"]). The number of elements returned must match the number of inputs provided by the \textbf{GetNextInputsAndCorrectOutput} method. 

The \textbf{GetNextInputsAndCorrectOutput} must return a tuple consisting of a single training set datapoint: an array of inputs and a corresponding output. These could be synthesized or read from a file, depending on the user’s needs. Note that the array of inputs should not be allocated inside \textbf{GetNextInputsAndCorrectOutput}. Instead, \textbf{inputsToFill} array passed as a parameter should be reused. This is done to reduce memory fragmentation and garbage collection overheads. 

\subsection{Configuration Parameters for Custom Machine Learning Experiments: Subclassing MLSetup – optional overrides}\label{sec:ConfigExperimentOpt}

Additional methods may be optionally overridden to refine and optimize machine learning performance:

\begin{verbatim}
    public virtual double SolutionFoundASRThreshold => 1.0;
\end{verbatim}

Overriding the \textbf{SolutionFoundASRThreshold} property enables the user to define the level of accuracy that constitutes an acceptable solution. The default value of 1.0 corresponds to a “perfect” solution, while values below 1.0 define tolerance thresholds for approximate solutions (e.g., \textbf{SolutionFoundASRThreshold = 0.9999} represents a relatively high-accuracy approximation). 

\begin{verbatim}
    public virtual int TargetColonySize(int generation)
\end{verbatim}

Overriding the \textbf{TargetColonySize} method enables the user to dynamically control the target population size as a function of the generation number. The default implementation (shown below) employs a periodic adaptive strategy: at initialization and at every 1,500 generations, the target colony size follows a staged reduction sequence—5 million for 5 generations, 3 million for 5 generations, 1 million for 5 generations, 500,000 for 5 generations—before returning to the baseline target of 250,000 individuals.  Note that the generation counter is reset upon colony reinitialization, as determined by the \textbf{TotalBirthsToResetColonyIfNoProgress} property (described later in this section).

\begin{verbatim}
    public virtual int TargetColonySize(int generation)
    {
        var remainder = generation % 1500;
        if (remainder < 5) return 5_000_000;
        if (remainder < 10) return 3_000_000;
        if (remainder < 15) return 1_000_000;
        if (remainder < 20) return 500_000;
        return 250_000;
    }

\end{verbatim}

\begin{verbatim}
    public virtual OpEnum[] GetAllowedOperations()
\end{verbatim}

Overriding the \textbf{GetAllowedOperations} method enables the user to restrict the operator set available for symbolic regression. For example, the implementation below excludes the sine and cosine functions from the search space: 

\begin{verbatim}
    public override OpEnum[] GetAllowedOperations() => 
    base.GetAllowedOperations().Where(x => x != OpEnum.Sin 
    && x != OpEnum.Cos).ToArray();
\end{verbatim}

Note that excluding the Copy and Paste pair is possible but is more involved and is not discussed in this document.

\begin{verbatim}
    public virtual uint ExperimentsPerGeneration => 512;
\end{verbatim}

Overriding the \textbf{ExperimentsPerGeneration} property modifies the number of experiments (i.e., “evolutionary runs”) executed per generation per organism. Since this parameter defines the thread count per GPU kernel block, optimal performance requires values divisible by 32 (i.e., CUDA warp size). Hardware constraints impose an upper limit of 1,024 threads per block. For point-to-point fitness evaluation, values of up to 1,024 are permissible; however, when employing correlation-based fitness functions, a maximum of 512 is recommended for computational efficiency. The default value of 512 provides optimal performance across most experimental scenarios. 

\begin{verbatim}
    public virtual long 
    TotalBirthsToResetColonyIfNoProgress => 25_000_000_000;
\end{verbatim}

Beagle implements a stagnation mitigation mechanism to escape local optima. When no fitness improvement is observed for a specified number of generations, the population undergoes complete reinitialization while preserving the “most accurate ever solution” discovered up to that point in the current run. Overriding the \textbf{TotalBirthsToResetColonyIfNoProgress} property modifies the generation threshold that triggers a population reset under stagnation conditions. This property can be set to -1 to prevent population resets.

\begin{verbatim}
    public virtual bool KeepOptimizingAfterSolutionFound => false;
\end{verbatim}
The \textbf{KeepOptimizingAfterSolutionFound} property determines whether the evolutionary search continues after identifying an acceptable solution (as defined by the \textbf{SolutionFoundASRThreshold} property). Continued optimization may yield shorter (i.e., more parsimonious) and/or more accurate solutions.

\begin{verbatim}
    protected virtual int ScriptLengthTaxRateInternal => 
    BConfig.MaxScore * (int)ExperimentsPerGeneration / 85;
\end{verbatim}

The \textbf{ScriptLengthTaxRateInternal} property controls the magnitude of parsimony pressure applied during evolution. This parameter is conventionally expressed as a ratio of the maximum achievable fitness and the number of experiments per generation. In the default implementation above, a solution of length 85 achieving perfect accuracy incurs a 100\% fitness penalty due to its length.

\subsection{Configuration Parameters for Custom Machine Learning Experiments: Configuration and Execution of ML Training in Beagle}\label{sec:ConfigExec}

The execution of machine learning training within the Beagle framework requires the specification of two primary components: (1) a subclass derived from \textbf{MLSetup}, and (2) a fitness function implemented as a struct implementing the \textbf{IFitFunc} interface. The fitness function must be defined as a struct rather than a class to satisfy the constraint of GPU kernel parameter passing, as reference types cannot be directly transferred to GPU computation units. 

The Beagle framework provides flexibility in fitness function selection, accommodating any user-defined fitness function that performs point-by-point comparisons between model-evaluated values and target values present in the training dataset. Beyond these custom implementations, Beagle incorporates a correlation-based fitness function, which has been previously demonstrated to enhance symbolic regression performance \cite{haut2026gpu}; this type of fitness function extended the methodologies established in prior research \cite{Haut2023}.

Training execution can be initiated through the following code in the Program.cs file of the Run project:

\begin{verbatim}
    using var mlEngine = new MLEngine<QuadraticEq, 
     CorrelationFitFunc>(forceCPUAccelerator: false);
\end{verbatim}

This instantiation configures the ML engine to utilize the \textbf{QuadraticEq} \textbf{MLSetup} with \textbf{CorrelationFitFunc} as the fitness evaluation function. The \textbf{forceCPUAccelerator} argument represents a boolean flag intended for kernel debugging purposes 

\subsection{Configuration Parameters for Custom Machine Learning Experiments: Fitness Functions}\label{sec:ConfigFit}

The framework provides several pre-implemented fitness functions available for immediate use. The \textbf{CorrelationFitFunc} operates as a correlation-based metric (illustrated in Equations \ref{eq:r} and \ref{eq:score}), while four standard point-to-point fitness functions are available: \textbf{StdFitFunc}, \textbf{StdSquareFitFunc}, \textbf{StdCubeFitFunc}, and \textbf{StdHyperFitFunc}. These standard functions share fundamentally similar computational logic (as illustrated in Equation \ref{eq:BeagleFit} for \textbf{StdFitFunc}), with the latter three variants applying exponential transformations to the base \textbf{StdFitFunc} result—specifically raising the output to powers of 2, 3, and 8, respectively. These power transformations serve to amplify fitness discrimination, thereby reducing the ASR for solutions that approach, but do not fully achieve, optimal fitness.


\begin{equation}\label{eq:r}
    r = \frac{\sum_{i=1}^{N} (y_i - \bar{y}) \, (f(x_i) - \overline{f(x)})}
{\sqrt{\sum_{i=1}^{N} (y_i - \bar{y})^2 \; \sum_{i=1}^{N} (f(x_i) - \overline{f(x)})^2}}
\end{equation}


\begin{equation}\label{eq:score}
    \text{score} = M \, r^4 \left( N - (c_1 + c_2) \right) - M (c_1 - c_2)
\end{equation}

Where $N$ is the number of data points, $M$ is the max reward/punishment; $y_i$ and $\hat{y_i}$  are target output and output from the model, respectively; $c1$ is the number of valid/invalid pairs; and $c2$ is the number of invalid/invalid pairs

\begin{equation}\label{eq:BeagleFit}
\text{Fitness}(\hat{y_i},y_i)
=
{\small
\begin{cases}
M,
& \max(|\hat{y_i}|,|y_i|) = 0, \\[0.8em]

\operatorname{round}\!\left(
M \cdot 
\frac{\min(|\hat{y_i}|,|y_i|)}{\max(|\hat{y_i}|,|y_i|)}
- \frac{M}{11}
\right),
& \hat{y_i}y_i < 0, \\[1.2em]

\operatorname{round}\!\left(
M \cdot 
\frac{\min(|\hat{y_i}|,|y_i|)}{\max(|\hat{y_i}|,|y_i|)}
\right),
& \hat{y_i}y_i \ge 0 .
\end{cases}
}
\end{equation}


Where $N$ is the number of data points, $M$ is the max reward/punishment; $y_i$ and $\hat{y_i}$  are target output and output from the model, respectively. 

\subsection{Configuration Parameters for Custom Machine Learning Experiments: Creating Custom Fit Functions}\label{sec:ConfigCustFit}

As stated in Section \ref{sec:ConfigExec}, a fitness function must be a struct implementing the \textbf{IFitFunc} interface. The interface definition is below:

\begin{verbatim}
    public interface IFitFunc
{
    bool UseCorrelationFit { get; }
    int FitFunction(float output, float correctOutput);
    int FitFunctionIfInvalid(bool 
     isOutputValid, bool isCorrectOutputValid);
}

\end{verbatim}

The \textbf{UseCorrelationFit} property is a boolean flag that dictates whether the fitness function should employ correlation-based evaluation logic that is hard-coded within the GPU kernel. When \textbf{UseCorrelationFit} is set to true, the remaining three methods defined in the \textbf{IFitFunc} interface are ignored during execution, and their implementations are not invoked by the training pipeline.

\begin{verbatim}
int FitFunction(float output, float correctOutput);
\end{verbatim}

The \textbf{FitFunction} method performs the principal fitness evaluation operation by comparing the model-generated output (i.e., \textbf{output} argument) against the target value (i.e., \textbf{correctOutput} argument) and computing an integer reward or penalty metric. This method is invoked exclusively under the condition that both \textbf{output} and \textbf{correctOutput} arguments constitute valid numerical values.

\begin{verbatim}
int FitFunctionIfInvalid(bool isOutputValid, 
  bool isCorrectOutputValid);
\end{verbatim}

In circumstances where either or both of the output values is/are not valid numbers, the \textbf{FitFunctionIfInvalid} method is invoked to determine the appropriate reward or penalty assignment for the model. This method accepts two boolean arguments that indicate the validity status of the model output and target output, respectively. By providing a dedicated pathway for handling invalid numerical states (such as NaN, infinity, or other undefined values), this method ensures robust and comprehensive fitness evaluation across all possible computational scenarios, including edge cases that arise during symbolic regression operations.

\subsection{Configuration Parameters for Custom Machine Learning Experiments: Useful Utilities}\label{sec:ConfigUtils}

The Utils directory under the Beagle directory in the BeagleLib project contains three useful command-line utilities: 

1.   \textbf{ExecuteGenome} – allows user to execute a genome with some input data;

2.	\textbf{Genome2Latex} – allows user to convert a genome to a LaTeX formula; 

3.	\textbf{MLSetup2File} – allows user to generate a CSV or JSON file with synthetic data based on a specific MLSetup.

4. \textbf{RunFullFeynman100Suite} - allows users to run the full Feynman symbolic regression benchmark. Runs all 100 equations to determine which equations are typically solved (at least 5/10) and occasionally solved (at least 1/10) in 10 independent 10-minute runs. 

\section{Conclusion}

This paper discusses Beagle, an open-source symbolic regression framework developed by Noblis, a non-profit science and technology organization based in the Washington, D.C. metropolitan area. A distinguishing characteristic of Beagle is its capacity to run populations comprising millions of individuals within computationally feasible timeframes. This performance enhancement relative to prior state of the art was realized  by leveraging GPU computational capabilities and the optimization of CPU-based components, including eliminating garbage collection overhead and maximizing concurrent execution across all available CPU cores. Consequently, Beagle demonstrated substantial performance advantages over traditional CPU-based genetic programming frameworks \cite{haut2026gpu}.

Beagle was conceived and developed independently of the broader evolutionary computing research community. During the framework’s development, we encountered and addressed numerous technical challenges through approaches that, in some cases (e.g., Linear Genetic Programming) converged upon existing methodologies through independent derivation, while in other cases (e.g., Monte Carlo Ranking selection) resulted in novel contributions to the field.

In the interest of benefiting the genetic programming community, Noblis has released Beagle as fully open-source software under the Apache 2.0 license, enabling researchers to examine and modify the source code. We believe that, in the future, GPUs may achieve an impact on evolutionary computing comparable to the impact they had on neural networks over the last decade. We anticipate that the genetic programming community will adopt this framework both for addressing practical symbolic regression problems and as a platform for future research, potentially yielding new insights, theoretical developments, and methodological advancements in genetic programming. We welcome collaborative engagement and community contributions to the codebase through pull requests. Inquiries and discussions regarding Beagle may be directed to Ilya Basin (ilya.basin@noblis.org) or Nathan Haut (nathaniel.haut@noblis.org).

\begin{credits}
\subsubsection{\ackname}    The team acknowledges Noblis, Inc. for supporting the development of the Beagle framework. This work was supported in part by Michigan State University through computational resources provided by the Institute for Cyber-Enabled Research.

\end{credits}
%
%
%
%
\clearpage
\bibliographystyle{abbrv}
\bibliography{references}

\end{document}